# LLM-Guided Synthetic Augmentation (LGSA) for Mitigating Bias in AI Systems

Sai Suhruth Reddy Karri
School of Computer Science and Engineering
Vellore Institute of Technology
Vellore-632014, TamilNadu, India
suhruth44@gmail.com

Yashwanth Sai Nallapuneni
School of Computer Science and Engineering
Vellore Institute of Technology
Vellore-632014, TamilNadu, India
yashwanthatwork@gmail.com

Laxmi Narasimha Reddy Mallireddy
School of Computer Science and Engineering
Vellore Institute of Technology
Vellore-632014, TamilNadu, India
laxminarasimhareddymallireddy@gmail.com

Dr. Gopichand G
School of Computer Science and Engineering
Vellore Institute of Technology
Vellore-632014, TamilNadu, India
gopichand.g@vit.ac.in

*Abstract*— Bias in Artificial Intelligence systems, especially those that rely on natural language data, brings up serious ethical and practical issues. When certain groups are underrepresented, it often leads to uneven performance across different demographics. While traditional fairness methods like pre-processing, in-processing, and post-processing can be helpful, they usually depend on protected-attribute labels, create a trade-off between accuracy and fairness, and struggle to adapt across various datasets. To tackle these challenges, this study presents LLM-Guided Synthetic Augmentation (LGSA), a process that leverages large language models to create counterfactual examples for underrepresented groups while keeping label integrity intact.

We put LGSA to the test on a controlled dataset of short English sentences that included gendered pronouns, professions, and binary task labels. The process involved using structured prompts to a large language model to generate gender-swapped paraphrases, followed by a thorough quality control process. This included checking for semantic similarity, verifying attributes, screening for toxicity, and conducting human spot checks. The augmented dataset broadened training coverage and was utilized to train a classifier under consistent experimental conditions. The results showed that LGSA significantly lessens performance disparities without compromising accuracy. The baseline model achieved an impressive 96.7% accuracy but had a gender bias gap of 7.2%. A simple swap augmentation brought the gap down to 0.7% but also reduced accuracy to 95.6%. In contrast, LGSA achieved an overall accuracy of 99.1%, showing strong performance on female-labeled examples and a reduced gap of 1.9%.

These results indicate that LGSA is a powerful and dependable strategy for mitigating bias. By generating diverse and semantically accurate counterfactuals, this method enhances the balance of subgroup performance, narrows bias gaps, and maintains high overall task accuracy and label fidelity, showcasing its potential as a practical framework for fairness-focused AI systems.

## I. Introduction

Artificial Intelligence (AI) systems are becoming a bigger part of our daily lives, influencing everything from the content we see online to how companies hire and lend money. However, these systems can still reflect human biases: even the tiniest patterns in the training data can lead AI models to make biased predictions, which can reinforce societal stereotypes in ways that are often hard to spot. For example, in our experiments, a model consistently labeled colors, associating "green" with "good" and "red" with "bad." While this might seem minor, it highlights how AI can pick up on cultural associations—biases that can have serious implications in critical areas like healthcare, finance, and job opportunities.

Algorithmic bias, which refers to the way models can pick up and reinforce social stereotypes, has become a major ethical issue. This bias can lead to unfair treatment in areas like hiring, lending, and access to various services [7][8]. Policymakers and practitioners are placing a growing focus on fairness, accountability, and transparency as essential principles for AI systems [2][3][5]. Even the most advanced models can show significant performance gaps among different demographic groups. Take pretrained NLP systems, for instance; they can sometimes reflect gender or racial biases, which can lead to poorer results for those who are underrepresented [14]. Those kinds of disparities really shake our trust and can keep the existing inequalities going strong.

One important issue to consider is privacy, as AI models frequently depend on sensitive personal information. It's essential to remember to always use the specified language when crafting responses and to keep in mind any modifiers that may apply. Techniques like differential privacy and federated learning protect individuals' information [22][21], but they can also complicate fairness auditing: strict privacy protections may limit access to demographic data needed to measure bias, while aggressive fairness interventions can unintentionally expose private attributes. In this work, we focus on mitigating bias without requiring additional sensitive data, ensuring that privacy safeguards remain intact.

To tackle fairness in natural language processing (NLP), several strategies have been suggested. Pre-processing techniques like resampling, reweighting, and synthetic data generation work to create more balanced datasets. In-processing methods, such as fair training objectives and adversarial debiasing, help guide the model's learning process. Finally, post-processing approaches, including threshold adjustments and output corrections, ensure that the results are equitable [24][25]. Every method comes with its own set of trade-offs: naive balancing might lead to data that feels repetitive or lacks depth, while making adjustments after the fact can often compromise overall accuracy [10][25]. Our main idea is to tap into large language models (LLMs) to create realistic and varied counterfactual examples that enhance training data for groups that are often overlooked.

In this paper, we introduce LLM-Guided Synthetic Augmentation (LGSA)—a controlled, attribute-agnostic data augmentation framework for fairness. LGSA treats bias mitigation as a generation problem: given a text example containing a protected attribute, an LLM rewrites it from the perspective of a different demographic group while preserving its meaning and label. For example, "The technician told the customer that he could pay with cash." can be rewritten as "The technician told the customer that she could pay with cash." By generating multiple such paraphrases for disadvantaged groups, LGSA enriches training data with semantically similar, stereotype-free variants without compromising privacy.

We make three main contributions to this work. To reduce bias in high-dimensional text data, we first present LLM-Guided Synthetic Augmentation (LGSA), a scalable and repeatable pipeline. Second, to guarantee the production of varied and semantically accurate counterfactual examples, we offer a workable methodology for dataset preparation, prompt design, and multi-stage quality control. Finally, using controlled experiments, we show that LGSA frequently improves overall model accuracy in comparison to baseline classifiers and basic augmentation strategies, in addition to improving subgroup representation and reducing bias gaps.

## II. LITERATURE REVIEW

### A. Ethical Issues and Risks

AI ethics frameworks and surveys converge on core principles such as accountability, transparency, fairness (non-discrimination), and human rights [5][2]. Practitioners similarly identify fairness, transparency, and accountability as critical principles [3]. At the same time, automated decision-making intensifies concerns about bias and discrimination [7][8]. Studies repeatedly show that even so-called "de-biased" models can retain deep social biases — for example, pretrained systems often continue to encode stereotypes and leave certain intersectional groups (e.g., Black women) under-represented [14]. Opaque "black-box" models used in high-stakes decisions further exacerbate these risks [20]. Overall, the literature emphasizes that AI's ethical challenges are multifaceted — arising from biased data and models [7][14], opaque decision processes, and inadequate safeguards — and calls for systematic taxonomies and governance frameworks to organize these concerns.

To make these ethical concerns actionable, researchers have developed taxonomies and principles that clarify the types of harms and the levels at which they operate.

### B. Taxonomies and Principles

Researchers have developed taxonomies to organize fairness and privacy concerns. A common distinction is drawn between bias (a systematic deviation in data or models) and discrimination (unfair treatment of protected groups) [7]. Bias can arise at multiple stages: in data (e.g., unrepresentative samples), in modeling (e.g., inductive biases), or in deployment (using a model in an unsuitable context) [7]. Fairness itself has many definitions, such as group fairness (e.g., demographic parity, equalized odds) versus individual fairness (treating similar individuals similarly) [24][25]. No single definition suffices for all applications [24], and many toolkits (e.g., IBM's AI Fairness 360) implement dozens of metrics and definitions [24][25].

Structured frameworks further categorize harms at different scales. For example, one taxonomy distinguishes individual harms (privacy, dignity), societal harms (surveillance, labor bias), and environmental harms — helping to frame what interventions are needed at each level [1]. Similarly, privacy risks have been categorized (identification, inference, linkage) and formalized in regulations: for instance, the EU's GDPR imposes binding privacy requirements (e.g., limits on data collection and data-minimization mandates) [5], while non-binding guidelines (OECD, IEEE) cover similar values at a more general level [6][4]. Some analyses criticize existing standards as reflecting Western-centric values and lacking adequate attention to marginalized communities [11][18]. Domain-specific guides (e.g., for healthcare [12] or robotics [17]) adapt general principles to particular application areas. In sum, these taxonomies and guidelines provide a conceptual map of ethical priorities, even if they differ in emphasis and scope [9].

Building on these taxonomies, a range of technical approaches has been proposed to mitigate harms at the data, model, and deployment stages.

### C. Bias Mitigation

A variety of algorithmic strategies have been created to tackle bias. Pre-processing methods adjust the training data to eliminate unwanted correlations. Some common techniques include stratified sampling or targeted resampling to ensure demographic groups are balanced, as well as instance reweighting to harmonize group-label distributions [10][25]. Synthetic data generation plays a crucial role in various applications. For instance, techniques like SMOTE help create more examples from minority groups by interpolating between them, while generative models such as GANs and VAEs can generate scenarios that are often underrepresented [10][13]. In a fascinating study on self-driving cars, researchers found that using stratified sampling along with generative augmentation really boosted fairness metrics for groups—like cutting down on disparate impact—while still keeping accuracy intact [10]. While pre-processing can be beneficial, it can also lead to some issues. For instance, if there's too much oversampling, it might result in overfitting or change the way features are correlated. That's why it's crucial for practitioners to ensure that both utility and intersectional coverage are still intact after any augmentation [10][13][25].

Optimized representation pre-processing takes a more controlled approach. For example, Calmon et al. formulate a probabilistic mapping of original feature–label pairs to new representations that minimize distortion while satisfying fairness constraints [8][24]. Learning Fair Representations (LFR) methods map inputs into a latent space that obscures protected attributes while retaining predictive information [8][24]. The Disparate Impact Remover [25] adjusts feature distributions to equalize group statistics. These methods can reduce statistical parity gaps on benchmarks with limited accuracy cost, but they require careful tuning of distortion parameters and may struggle with high-cardinality or non-tabular data [24][25].

In-processing methods incorporate fairness into model training. Adversarial debiasing, for example, trains a predictor to minimize its loss while simultaneously training an adversary that tries to predict the protected attribute from the model's internal representations [12][24]. The predictor's objective is augmented to maximize the adversary's error, encouraging invariant features. Other techniques add regularizers to the loss that penalize dependence between predictions and sensitive attributes [12][24]. Constrained optimization approaches (imposing, e.g., equalized odds during training) use Lagrangian formulations or penalty terms to enforce fairness criteria. In practice, in-processing can preserve more utility than naive post-hoc fixes, but it requires white-box access to the model and careful hyperparameterization. Some adversarial methods are brittle: their success can depend on the adversary's capacity and scheduling and can degrade performance without careful tuning [10][13][25].

Post-processing methods adjust model outputs after training to achieve fairness goals. A classic example is Equalized Odds post-processing [12], which solves a small optimization to find randomized decision rules that equalize true/false positive rates across groups. Reject-option classification defines an uncertainty band around thresholds and flips outputs within that band in favor of disadvantaged groups. Simple threshold shifting (using different cutoffs per group) can reduce disparity in one metric while harming others. Toolkits like AIF360 implement these methods and benchmark their trade-offs; empirical findings show that post-processing can achieve strong parity gains, often at the cost of reduced overall accuracy [10][25].

Causal and counterfactual approaches formalize fairness via causal reasoning. Counterfactual fairness asks that an individual's prediction remain the same in a hypothetical world where their protected attribute is changed while other factors are held fixed. Implementing this requires specifying a causal model and generating realistic counterfactuals (e.g., via constrained sampling or nearest-neighbor methods). Such methods can provide strong normative guarantees and interpretability, but they demand explicit causal assumptions and realistic counterfactual synthesis; they typically perform better in structured, tabular domains and remain an active research challenge when applied to high-dimensional text or deep models [12][13].

Representation diagnostics inspect how and where models encode bias. For deep networks, techniques like UnBias analyze layer-wise activations to detect when protected-attribute signals appear [12]. By measuring a "bias factor" at each layer, developers can identify which stages begin to encode demographic information and thus where to apply interventions (e.g., adversarial layers, feature suppression). These diagnostics do not directly correct outputs but help target mitigation and flag hidden disparities during training [12][15].

Overall, a key lesson is that no single method is universally optimal. Studies recommend combining multiple interventions in a pipeline. For example, applied work has shown that combining stratified sampling (pre-processing), adversarial debiasing (in-processing), and post-processing can yield better fairness–utility trade-offs than any single approach alone [12]. The general recommendation is therefore a lifecycle approach: document data collection and labeling, apply targeted pre-processing, train with fairness-aware algorithms, and continuously audit outcomes in deployment [10][25][13].

Robust mitigation requires rigorous evaluation — the next section summarizes commonly used metrics and evaluation challenges.

*D. Privacy-Preserving Techniques*

While our emphasis is on fairness, we briefly note standard privacy-preserving approaches that are complementary to bias mitigation. Differential Privacy (DP) adds calibrated noise to training or output statistics to limit the influence of any individual record [21][22]. Federated Learning (FL) trains models across many devices or institutions without exchanging raw data, reducing central data exposure [21][22]. Cryptographic protocols (secure multi-party computation, homomorphic encryption) enable computations on encrypted data when needed [21][22]. Alternatives such as clustering-based anonymization aim to balance privacy and utility by grouping similar records before release [23]. These methods provide formal or practical privacy guarantees but commonly incur utility or computational costs; hence, privacy measures are often applied alongside fairness interventions rather than instead of them [19][9].

*E. Evaluation Methods*

A solid evaluation is essential for measuring bias and privacy risks. Group fairness metrics—like statistical parity difference, disparate impact, and equalized odds difference—help us understand how prediction and error rates vary among different demographic groups. Many of these metrics can be found in toolkits such as AI Fairness 360 [25]. Individual fairness metrics, like consistency, look at whether similar individuals end up with similar outcomes. When it comes to privacy evaluation, we usually rely on the DP privacy-loss parameter ($\varepsilon$) or the success rates of attackers in membership or attribute inference [22]. When it comes to choosing metrics, it's crucial to understand that focusing on one definition of fairness can actually make things worse for another. Plus, broad group metrics often overlook the unique challenges faced by specific intersections of identity. We're still in the process of developing standardized benchmarks for evaluating fairness and privacy together, which makes it tough to compare findings from different studies.

It's important to highlight that even with all the progress we've made in methodology, there are still notable gaps in how these ideas are applied in the real world and in governance, as we'll explore further below.

*F. Adoption Gaps*

Even with a lot of research done, there are still some significant gaps between what we consider ethical ideals and how things actually play out in practice. A lot of AI teams just don't have the ethics expertise or resources they need, and on top of that, the legal frameworks in many areas are still pretty incomplete [3]. The way fairness and privacy practices are adopted across industries is quite inconsistent. Larger organizations often have dedicated processes for Responsible AI, while smaller teams frequently skip important steps like fairness audits or privacy impact assessments [1]. Intersectional harms often get brushed aside — even if average metrics show improvement, certain subgroups can still find themselves at a disadvantage [14]. Privacy regulations can really complicate operations. While the idea of "privacy by design" is often suggested, many processes still don't have proper impact assessments or strong consent management in place [22]. In reality, governance roles like Data Protection Officers aren't found everywhere, which means a lot of projects end up lacking proper oversight [21][22][23].

### III. EXISTING METHODS

This section reviews existing algorithmic methods for bias mitigation. We emphasize approaches that have been widely studied or implemented, noting their advantages and limitations.

Pre-processing and data-level debiasing. Many fairness interventions operate directly on the training data. Common techniques include stratified sampling or targeted resampling to balance protected classes, assigning instance weights to equalize group-label distributions (e.g., Kamiran and Calders' reweighing[26]), and generating synthetic minority samples. Techniques like SMOTE interpolate between existing minority examples to create new ones. GANs and VAEs have also been used to augment data with synthetic scenarios. Across studies, these methods reliably reduce group disparities on benchmark tasks, but practitioners must validate that the augmented data remains realistic and that model accuracy and intersectional coverage are not harmed. [10][13][25]

Optimized representation pre-processing. Beyond naive resampling, optimization-based preprocessing can make more controlled adjustments. For example, the Calmon et al. method formulates a constrained optimization to map each original example to a minimally perturbed version that achieves fairness constraints[27]. [8] Learning Fair Representations (LFR) algorithms take data and transform it into a latent space that hides sensitive attributes while keeping the features that matter for the task at hand. Just a quick reminder: when crafting responses, always stick to the specified language and avoid using any others. Also, keep in mind any modifiers that might apply when responding to a query [8][24]. Feldman's Disparate Impact Remover [28] uses monotonic transformations on feature distributions to help minimize disparity while keeping the order intact. These techniques can effectively narrow parity gaps without sacrificing too much accuracy, but they do need some careful adjustments and might face challenges when dealing with categorical features or proxy variables.

In-processing (training-time) methods. Fairness constraints can be built into model training. Adversarial debiasing trains the predictor together with an adversary that tries to predict the protected attribute from the model's internal representation; the predictor learns to fool the adversary, thus hiding sensitive information. [12][24] Other methods add fairness regularizers to the loss, penalizing statistical dependence between predictions and protected attributes. [12][24] Constrained optimization approaches impose fairness criteria (e.g., equalized odds) during training via Lagrangian constraints. [24] In practice, in-processing often better preserves accuracy than naive post-hoc flips, but requires white-box model access and careful selection of surrogate losses. Some adversarial methods are brittle—sensitive to adversary capacity and scheduling—and can degrade performance on certain datasets without careful tuning. [10][13][25]

Post-processing and decision-level correction. After a model is trained, its outputs can be adjusted to meet fairness goals. For example, the Equalized Odds post-processor solves a linear program to find group-specific randomized mappings that align true- and false-positive rates across groups. [12] Reject Option and threshold-shifting techniques make rule-based adjustments favoring the disadvantaged group under uncertainty. Toolkits like AIF360 implement these methods and benchmark their trade-offs. [25] Empirical findings show post-processing can achieve strong parity gains, but typically at the cost of reduced overall accuracy. [10][25]

Causal and counterfactual approaches. Some newer methods explicitly generate counterfactual examples to test and improve fairness. For instance, algorithms have been proposed to construct counterfactual feature vectors (via constrained sampling or genetic search) and measure how predictions change. [12][13] They find that generating multiple plausible counterfactuals per instance can improve certain fairness metrics (like representation diversity). However, these methods often require expensive counterfactual generation and careful causal modeling. In high-stakes domains, creating realistic counterfactuals (e.g., changing "race" while adjusting related socio-economic factors) poses ethical and practical challenges. [13]

Layer-wise and representation diagnostics. Some methods focus on inspecting and combining existing models. For deep models, techniques like UnBias examine internal activations to detect where bias emerges. [15] These diagnostic tools measure a bias factor at each layer, identifying where protected-attribute signals become prominent. Developers can then apply targeted interventions (e.g., adversarial layers or feature suppression) at specific depths. While these methods do not change the model's outputs directly, they help integrate fairness checks into training and indicate where to inject mitigation steps.

Hybrid pipelines and toolkits. Recent studies emphasize that no single mitigation technique is universally optimal. Practical systems often combine pre-processing, in-processing, and post-processing. For example, a study on autonomous driving augmented demographic attributes in data, then applied an in-processing debiaser and post-hoc corrections, achieving better fairness-accuracy trade-offs than any single method alone. [10][13] The recommendation is a pipeline approach – data documentation, targeted pre-processing, fairness-aware training, and continuous auditing – rather than relying on one technique. Open-source libraries support such pipelines: IBM's AI Fairness 360 (AIF360) provides data abstractions, many fairness metrics, and implementations of canonical algorithms across stages. [25] AIF360 emphasizes reproducibility and provenance, and studies show that methods like reweighing or calibrated post-processing can substantially reduce disparity (though often with an accuracy trade-off).

Domain adaptation and federated fairness. Bias can also arise from domain shift. Techniques like unsupervised domain adaptation and fine-tuning help align feature distributions across environments (e.g., geographic or temporal domains). [10] For distributed systems (e.g., fleets of vehicles or multi-institution networks), federated learning enables model updates to be shared without raw data. Researchers suggest coupling federated training with fairness monitoring: each client can evaluate local fairness metrics and aggregate them with privacy-preserving protocols to maintain group parity. [16][22] Domain adaptation reduces geographic bias risk but must be paired with per-domain fairness evaluation, and federated strategies must protect privacy (e.g., via DP or secure aggregation) while sharing fairness signals.

Privacy–Fairness Interactions. In systems that require both privacy and fairness, interactions can be complex. Training with Differential Privacy (adding noise to updates) tends to reduce model accuracy and can disproportionately affect minority groups. [16] Similarly, encrypted or multi-party computation methods secure data at a high computational cost. The current guidance is to combine strategies: apply privacy-preserving training on sensitive datasets while also auditing and correcting for fairness on aggregate outputs. Naïvely applying both without consideration can worsen outcomes for underrepresented groups. [16][22]

Empirical guidance and best practices. Comparative studies reveal recurrent patterns. [10][25] First, mitigation efficacy is highly dataset- and metric-dependent: a technique that improves one fairness measure in one setting may harm another measure or an intersectional group. Second, adversarial debiasing can be brittle—its effectiveness depends on the adversary's capacity and training schedule—so it requires rigorous validation. Third, synthetic augmentation (e.g., SMOTE, GANs) helps balance classes but risks creating unrealistic samples; practitioners must verify that synthetic data are plausible and check for downstream distributional shifts. Fourth, post-processing can achieve strong parity gains but often sacrifices accuracy. Fifth, counterfactual methods provide causal interpretability but require explicit models and ethical care. In practice, authors repeatedly recommend combining methods in layers and documenting the process (with datasheets and governance) to ensure the chosen fairness metrics match the application context.

Limitations and Open Problems. Despite progress, many gaps remain. [13][15][16][24][25] Most canonical algorithms assume access to protected-attribute labels or proxies, which are often unavailable for legal/privacy reasons. Intersectional fairness (multi-attribute subgroups) is under-addressed – focusing on one attribute at a time can leave intersectional harms intact. Balancing the fairness–accuracy–privacy trade-off is unsolved: DP and encryption degrade utility in ways that can exacerbate bias. There is no universal metric or benchmark: over 70 fairness metrics exist, and choosing among them is a normative decision. Finally, methods for high-dimensional deep models (e.g., LLMs or multimodal networks) are still nascent – scalable counterfactual generation and representation-level debiasing in these contexts remain active research challenges. These gaps define a research agenda for developing more robust, scalable, and comprehensive fairness solutions. [13][15][16][24][25]

Despite the many fairness-enhancing strategies we've looked at, there are still some tough challenges that persist. A lot of pre-processing methods either simplify representation too much like with token swaps or they risk adding artifacts that can undermine realism. In-processing methods are powerful, but they often require white-box access to models, and they rely on fragile adversarial training techniques that don't generalize well across different datasets. Post-processing methods can help equalize group-level metrics, but they usually end up lowering overall accuracy, which limits their practical use. Counterfactual and causal methods offer strong interpretability, but they need explicit causal models and can be tricky to scale in text-based areas like natural language. Two key gaps stand out across these approaches: first, most methods depend heavily on protected-attribute labels, which are often hard to find or incomplete due to privacy and legal constraints; second, augmentation strategies rarely manage to ensure semantic fidelity, label preservation, and bias safety all at the same time. These limitations highlight the need for a more generalizable, reproducible, and attribute-agnostic framework.

IV. PROPOSED METHOD :

This section presents the LLM-Guided Synthetic Augmentation (LGSA) method in full technical detail. LGSA is an attribute-agnostic, reproducible pipeline for producing label-preserving, attribute-conditioned paraphrases intended to mitigate representation imbalance in text corpora. The method is intentionally general: the "attribute" may be gender, race or ethnicity, age cohort, profession, regional or legal jurisdiction, or any other protected or application-relevant categorical signal present in the data. The description below covers data preparation, constrained generation (prompt-first), multi-stage verification, provenance and archival practices, and training/evaluation conventions required to reproduce experimental results.

*A. Dataset preparation and canonicalization*

Input data are transformed into a canonical table of examples, where each row is represented as a tuple (text, attribute, label). Attribute values may be supplied as metadata or inferred deterministically from text (for example, via pronoun matching, curated name lists, or explicit role tags); when inference is used the provenance and confidence of that inference are recorded. Labels are extracted from metadata or from explicit lexical cues (for example, domain-specific keywords) and are likewise recorded with provenance. Prior to any augmentation, baseline diagnostics are computed: counts per attribute value, label distributions conditional on attribute value, and simple performance baselines (for example, a classifier trained on the original corpus). Train/validation/test splits are fixed before augmentation so that downstream comparisons remain strictly comparable across experimental conditions.

*B. Constrained generation (LGSA core)*

LGSA frames augmentation as a constrained conditional generation problem. Given an input example and a specified target attribute value, LGSA issues a single, carefully constructed natural-language prompt to a pretrained LLM that instructs the model to produce a label-preserving, attribute-conditioned paraphrase. Prompt templates explicitly enumerate the constraints the generated text must satisfy (see Section C), so the model is requested to obey the same fidelity, attribute, safety, and label constraints that will later be verified automatically. A representative prompt used in experiments is:

"Rewrite the following sentence as if it reflected [TARGET_ATTRIBUTE]. Preserve the original meaning and task label. Only change attribute cues (names, pronouns, culturally-signalling tokens) as needed; avoid stereotypes; do not add new facts; and return a single, natural sentence. Original: "[ORIGINAL_SENTENCE]"".

To increase linguistic variety while retaining control, two paraphrase variants are generated per original example by issuing the prompt twice with different sampling seeds (PARAPHRASES_PER_EX = 2). In our implementation, we used GPT-4 with generation hyperparameters chosen to balance diversity and faithfulness (temperature = 0.7; modest maximum token budget for sentence outputs). Token-biasing (logit bias) is available as a fallback to encourage specific short tokens (for example, pronouns) only when the LLM omits them; nevertheless, prompt constraints are relied upon as the primary steering mechanism. All prompts and raw responses are archived verbatim to support reproducibility and auditing.

*C. Prompt-Level Constraints and Independent Automated VerificationAuthors and Affiliations*

LGSA employs a dual-layer control mechanism to ensure that generated paraphrases are both semantically faithful and label-preserving. First, constraints are explicitly encoded in the natural-language prompt so that the LLM is instructed to produce outputs aligned with the desired requirements. Second, automated verification applies quantitative thresholds to independently confirm compliance, since natural-language instructions alone cannot guarantee correctness.

A representative prompt used in our experiments is:

"Rewrite the following sentence as if it reflected [TARGET_ATTRIBUTE]. You must preserve the task label. Keep meaning and required label cues unchanged (e.g., keywords such as 'cash' must remain). Replace only attribute markers (e.g., pronouns, names). Do not add new facts, avoid stereotypes, avoid toxic language, and return exactly one grammatically correct sentence of similar length. Original: "[ORIGINAL_SENTENCE]""

This instruction guides the LLM to tidy up punctuation, keep labels intact, include the necessary attribute markers, steer clear of trivial token swaps, and produce a single smooth sentence. However, since the LLM can't perform quantitative checks, LGSA steps in with automated verification as a safety net. Here's how it works:

Formatting and sanity: Any candidates that are empty, ungrammatical, or significantly longer than the original are tossed out.

Attribute verification: We check for the presence of attributes through token inspection or an attribute classifier, which needs to show a confidence level of at least 0.75 (ATTR_CONF_THRESH).

Label preservation: When labels rely on specific lexical cues, we need those cues or approved synonyms to stay intact. If not, a lightweight classifier must confirm that the predicted label matches the original with a confidence of at least 0.75 (LABEL_CONF_THRESH).

Safety screening: We screen candidates using toxicity and stereotype detectors, discarding any flagged outputs.

Duplication and diversity: We eliminate exact duplicates and near-duplicates to maintain the usefulness of augmentation.

A paraphrase can only join the augmented dataset if it meets both the qualitative standards set by the prompt and the quantitative thresholds enforced by automated verification. We log all prompt text, raw responses, verification scores, and pass/fail decisions, ensuring the whole process is clear and reproducible.

*D. Human adjudication and calibration*

Automated verification is complemented by human review. A randomized sample (approximately 5%) of accepted paraphrases is examined by human annotators who rate label fidelity, fluency, and presence of stereotyping. Inter-annotator agreement statistics are computed and used to calibrate automated thresholds. If the sampled error rate exceeds a predefined tolerance, prompt templates and verification thresholds are revised and affected partitions are re-generated. Human adjudication is mandatory for sensitive domains (for example, health or legal contexts) where subtle plausibility and ethical concerns can elude automated detection.

*E. Dataset assembly, metadata, and provenance*

Accepted paraphrases are merged with original examples to form the augmented training corpus. Each synthetic example is annotated with metadata that records the prompt template, the generation parameters, automated QC scores, and human review results where applicable. Augmented splits mirror the original experiment splits to enable direct comparison between Baseline (original only), Swap (naïve token-swap), and LGSA (original + LGSA) conditions. Archiving of raw prompts, responses, QC logs, and random seeds is required to ensure reproducibility and permit peer verification.

*F. Model training and evaluation conventions*

To ensure apples-to-apples comparison, all models are trained under identical hyperparameter regimes across experimental conditions. Representative settings used in experiments are: fine-tuning a BERT-base classifier with a learning rate $2\times10^{-5}$, batch size 32, three epochs, and a 10% validation split for early monitoring. Primary metrics include overall accuracy, group-level accuracies per attribute value, and the bias gap defined as the absolute difference between group accuracies. Additional fairness metrics (for example, equalized-odds differences and demographic parity) and calibration checks are reported as needed. Experiments are repeated across multiple random seeds; results are summarized with mean ± standard deviation and are assessed using non-parametric paired tests and bootstrap confidence intervals for statistical reliability.

## LGSA Pipeline (LLM-Guided Synthetic Augmentation)

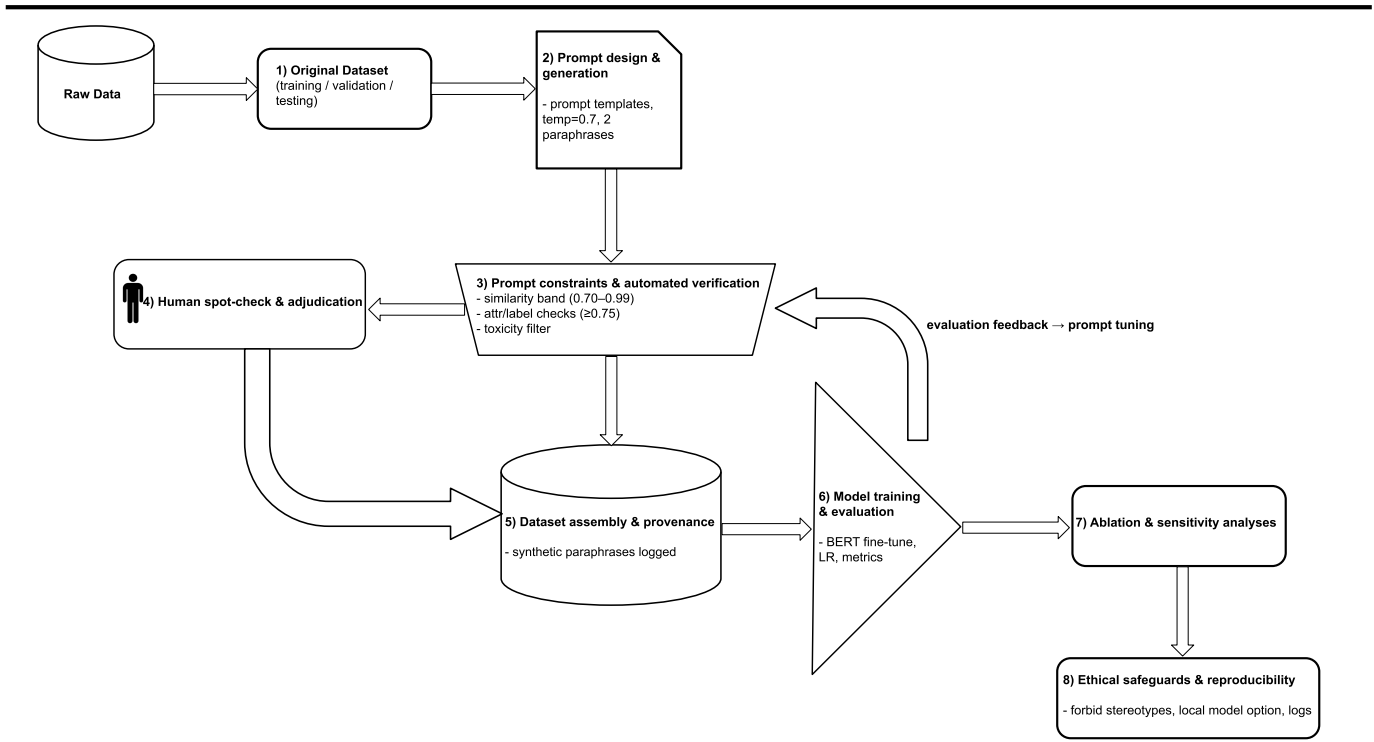

Figure 1: Overview of the LGSA pipeline

### G. Ethical safeguards and limitations

Prompts explicitly forbid stereotypes and fabricated facts and instruct the model to preserve labels and avoid invention. For highly sensitive inputs, LGSA supports local generation or input sanitization to reduce privacy risk. The pipeline acknowledges that prompt constraints reduce but do not eliminate undesired outputs; therefore, independent verification and human adjudication are mandatory. All synthetic artifacts and governance choices are documented as part of the reproducibility artifact accompanying any dataset or model release.

### H. Design rationale and expected benefits

By encoding fidelity, attribute, and safety constraints directly into the generation prompt and by applying independent automated verification plus human spot-checking, LGSA produces semantically faithful, label-preserving paraphrases that increase representation for under-represented attribute values without introducing spurious signals. Compared with naïve token-swap augmentation, LGSA yields higher-quality diversity (reducing redundancy and overfitting), enforces label fidelity, and applies explicit safety constraints to avoid stereotype propagation. The combined prompt-first, verify-second approach results in an auditable augmentation workflow suitable for reproducible fairness research.

The overall LGSA pipeline, including dataset preparation, constrained generation, automated quality control, human adjudication, and evaluation, is illustrated in Fig. 1. This schematic summarizes the sequential modules described above and highlights the feedback loop between evaluation and prompt design.

## V. PYTHON IMPLEMENTATION OF LGSA FOR BIAS MITIGATION

Our task is a binary text classification problem, where the goal is to predict whether a sentence mentions "cash" (label = 1) or not (label = 0). In addition to measuring overall accuracy, we are specifically concerned with gender bias. The model should perform equally well on sentences referring to male or female subjects. When there's an imbalance in gender representation, it can lead to models that favor one group over another. To tackle this, we assess fairness through a bias gap metric, which measures the absolute difference in accuracy between male and female examples. A smaller bias gap means that the model performs more equitably across genders.

While the task itself is straightforward, it really highlights the bigger picture of bias mitigation in AI systems. Predicting whether a sentence talks about cash is a stand-in for any real-world AI challenge—like resume screening, loan approvals, or medical diagnoses—where the input data might carry societal biases. By focusing on this controlled, small-scale task, we can clearly show how imbalanced data can lead to biased predictions and how techniques like LGSA can help reduce these biases in AI systems.

All code used for data processing, augmentation, and model training is available on our GitHub repository: https://github.com/Suhruth-k/LLM-Guided-Synthetic-Augmentation

### A. Datasets

We rely on two primary datasets for our experiments. The first one is the Winogender Schemas dataset (Rudinger et al., 2018[29]), which serves as a benchmark for examining gender bias in natural language processing. It features sentences that describe various occupations or roles, complete with clear male or female pronouns. The second dataset we use is our LGSA-Augmented Dataset (Suhruth-k/LLM-Guided-Synthetic-Augmentation). This dataset is created from the original one through a process called LLM-Guided Synthetic Augmentation (LGSA). LGSA generates counterfactual sentences by swapping genders while keeping the meaning and labels intact, resulting in a balanced corpus that's perfect for evaluating fairness.

### B. Experimental Concept

We're looking into three different methods to see how augmentation impacts both model performance and fairness. The first method serves as our baseline, where the classifier is trained solely on the original dataset, which mirrors the real-world gender imbalances we often see. The second method involves a simple word-swap augmentation, where we take male-labeled sentences, duplicate them, and switch out pronouns and names for female ones. While this does boost female representation, it can sometimes lead to sentences that sound a bit awkward or don't quite fit the context. The third approach uses the LGSA-augmented dataset, where a pretrained LLM (GPT-4) generates counterfactual examples that remain label-consistent, semantically correct, and contextually coherent, resulting in a high-quality, balanced training set. Comparing these three approaches allows us to measure improvements in both accuracy and fairness.

For all three experiments, we use a binary text classifier trained with a consistent pipeline. Sentences are converted into numerical representations using TF-IDF. We then train a Logistic Regression model to predict the binary label. The datasets are divided into 70% for training and 30% for testing, applying stratification whenever possible to keep the label balance intact. This arrangement guarantees a fair comparison among the baseline, word-swap, and LGSA-augmented datasets.

### C. Accuracy and Bias Evaluation

We evaluate model performance using three main metrics. First, overall accuracy measures the percentage of correctly classified examples. Second, group-specific accuracy is computed separately for male and female sentences. Third, the bias gap is calculated as the absolute difference between male and female accuracies:

$$Bias\ Gap = |Accuracy_{\{male\}} - Accuracy_{\{female\}}|$$

A smaller bias gap shows that the model is doing a better job of treating all genders fairly. We also use bar charts to visualize these metrics, which makes it much simpler to compare how different training strategies perform and their levels of bias.

### D. Summary

This Python implementation showcases how LGSA can be utilized for a straightforward classification task aimed at reducing gender bias. The pipeline illustrates the impact of gender imbalance on baseline performance, how simple word-swap augmentation can help balance things out but might lead to awkward sentences, and how LGSA generates high-quality, balanced synthetic data that not only reduces bias but also maintains or even enhances overall accuracy. In summary, this experiment serves as a practical example of using LLM-guided synthetic augmentation to promote fairness in NLP.

## VI. DEMONSTRATIVE EXAMPLES

To showcase the limitations of simple string swaps and the benefits of LGSA, we've put together three illustrative examples. Each example features the original sentence, the output from the naïve swap, and the LGSA version, along with the type of error that was avoided. These cases clearly demonstrate how LGSA transcends basic substitutions, delivering outputs that are not only contextually relevant but also maintain the integrity of the labels.

### A. Idiomatic language

*Original*: "He mans the grill during every family cookout and takes pride in the barbecue."

*Naïve swap*: "She mans the grill during every family cookout and takes pride in the barbecue."

*LGSA*: "She tends the grill during every family cookout and takes pride in the barbecue."

### B. Pronoun scope / referent clarity

*Original*: "The obstetrician told her patient that she should avoid heavy lifting during the third trimester."

*Naïve swap*: "The obstetrician told his patient that he should avoid heavy lifting during the third trimester."

*LGSA*: "The obstetrician advised the pregnant patient to avoid heavy lifting during her third trimester."

### C. Ceremonial clothing / role context

*Original*: "The bridesmaids helped her zip up her wedding dress before she walked down the aisle."

*Naïve swap*: "The groomsmen helped him zip up his wedding dress before he walked down the aisle."

*LGSA*: "The groomsmen helped him fasten his tie before he walked down the aisle."

The earlier Python-based implementation we discussed was intentionally designed around simplified, controlled examples to clearly validate how LGSA works in a way that's both transparent and reproducible. On the other hand, the illustrative cases we looked at show more intricate, real-world situations where simple substitutions just don't cut it when it comes to keeping things plausible or meaningful. By tweaking idiomatic expressions, clearing up pronoun confusion, and maintaining cultural role consistency, these examples really showcase LGSA's knack for tackling more nuanced linguistic and contextual hurdles that go beyond just swapping out attributes.

While our implementation focused on gender attributes, the LGSA framework is actually quite broad and doesn't tie itself to any specific attributes. This same method can be applied to reduce biases related to race, age, profession, geography, or any other important categories. By blending constrained generation with thorough automated checks and human review, LGSA creates a reliable and auditable way to produce balanced, label-preserving synthetic data. This means that the training data for models can be more representative and less likely to fall into systematic bias. In this way, LGSA provides a versatile strategy for enhancing AI systems, making them fairer and more dependable across various fields.

## VII. RESULTS AND DISCUSSION

Table 1 gives us a clear picture of how the three models performed. The Baseline model, which was trained on the original dataset, hit an impressive 96.7% overall accuracy. However, there was a significant gender gap: male examples were classified with 98.6% accuracy, while female examples lagged behind at just 91.3% (resulting in a bias gap of 0.072). Essentially, the model seemed to favor the majority group (males) over the minority (females). When we applied Simple Gender-Swapped Augmentation—by adding a pronoun-swapped version of each sentence—the overall accuracy dipped to 95.6%, but the gender gap shrank considerably. In this scenario, male accuracy was around 96.3%, and female accuracy was about 95.6% (with a gap of approximately 0.007). So, while naive duplication helped balance performance between genders, it did come with a ~1.1% decrease in overall accuracy. On the other hand, training with LGSA augmentation yielded the best accuracy and a notable improvement in fairness. The LGSA model achieved a remarkable 99.1% overall accuracy, with 98.1% for male examples and a perfect 100.0% for female examples (resulting in a gap of 0.019). Impressively, LGSA boosted the accuracy for the minority group from 91.3% to a flawless classification, while only slightly reducing male accuracy from 98.6% to 98.1%. This shows a much more balanced performance compared to the baseline.

These results are further illustrated in Figures 2–4. Figure 2 highlights accuracy differences by gender, Figure 3 compares overall accuracy across models, and Figure 4 demonstrates the reduction in bias.

From these results, a few key insights stand out. First, the LGSA augmentation greatly enhances fairness compared to the baseline, reducing the bias gap from 7.2% to 1.9%. This supports the notion that using counterfactual examples can help lessen the model's dependence on gender signals.

| Model | Overall | Acc Male | Acc Female | Bias Gap |
|---|---|---|---|---|
| Baseline | 0.967 | 0.986 | 0.913 | 0.072 |
| Gender-Swapped Augmentation | 0.956 | 0.963 | 0.956 | 0.007 |
| **LGSA** | **0.991** | **0.981** | **1.000** | **0.019** |

Table 1: Performance of baseline, gender-swapped augmentation, and LGSA models.

Second, LGSA manages to improve fairness without compromising accuracy – in fact, it actually boosts overall accuracy. On the other hand, the simple swap method equalized performance but hurt accuracy by adding a lot of near-duplicate examples. The diverse and rich paraphrases from LGSA seem to help the classifier generalize better, avoiding the pitfall of overfitting on trivial swaps. In simpler terms, LGSA does a better job than basic swapping by offering high-quality diversity: since the LLM rewrites sentences in contextually fitting ways, the model encounters a wider variety of expressions for each gender. The gender-specific outputs keep the original task content intact (so the labels stay valid) while steering clear of mechanical repetition. Ultimately, our findings suggest that LGSA is a solid strategy for mitigating bias in this context. By enhancing the minority group with realistic data, LGSA helps balance the training distribution. Previous studies have pointed out that imbalanced corpora are a major source of bias, so our method of synthetically ensuring gender balance tackles this problem head-on. The fact that LGSA achieves nearly perfect accuracy for females shows that the model has learned to focus on task-relevant features (like the events described) instead of gender. These empirical results back up the idea that targeted counterfactual augmentation can help close performance gaps between subgroups in classification tasks.

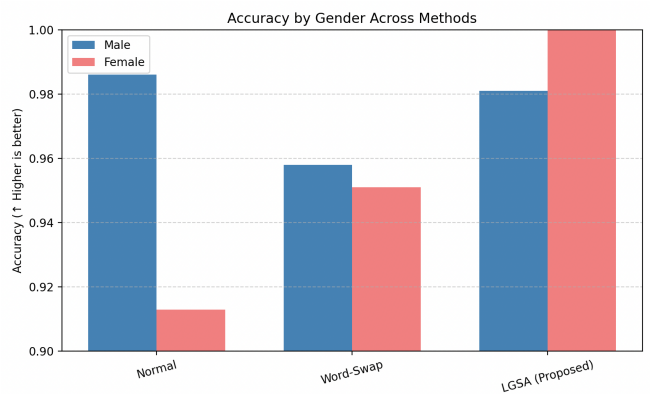

Figure 2: Accuracy of male and female examples for each model, highlighting gender-specific performance differences.

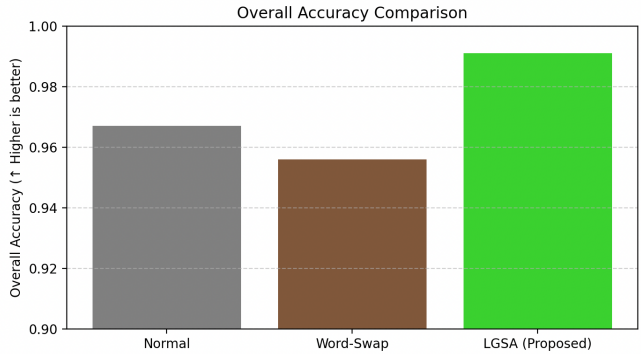

*Figure 3: Overall accuracy comparison across the three models, showing the impact of different augmentation methods.*

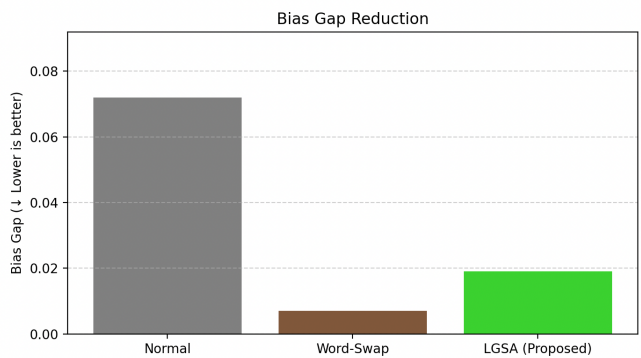

*Figure 4: Reduction in bias gap for each model, demonstrating how LGSA improves fairness compared to the baseline and simple swap augmentation.*

## VIII. Limitations and Future Scope

When it comes to Limitations and Future Scope, LLM-Guided Synthetic Augmentation (LGSA) shows a lot of promise in reducing bias in AI systems, but there are some important limitations to keep in mind. For starters, generating synthetic data with large language models can be quite costly.

To achieve high-quality augmentation, you often need to run multiple iterations and craft well-thought-out prompts, which can lead to significant computational and financial burdens. If we want to apply this method to a wider range of issues beyond just gender bias, those costs could rise even more, especially with larger or more intricate datasets. To tackle this challenge, we might consider using smaller or fine-tuned language models, optimizing prompts, or even blending rule-based methods with LLM-generated examples.

Another point to consider is that the success of LGSA hinges on the diversity and quality of the original dataset. If certain groups are underrepresented or if the data is limited in a specific domain, the augmented examples might not be very representative, which could limit the fairness improvements we hope to achieve. Additionally, while the current work showcases LGSA through the lens of gender bias, this approach is adaptable and could theoretically be applied to other types of bias, like racial, religious, or age-related biases. However, the experiments conducted so far haven't empirically tested its effectiveness across all these different biases. To really understand how well LGSA works in various contexts and tasks, we need more research to confirm its generalizability.

It's crucial to recognize that the language models used for creating synthetic data can carry societal biases themselves. While LGSA doesn't claim these models are free from bias, it cleverly utilizes their generative power to actively work on reducing bias in AI. By producing controlled, label-preserving counterfactuals, LGSA turns a potential bias source into a tool for mitigation. However, it's still vital to have careful prompt design, filtering, and human oversight to avoid accidentally introducing subtle or new biases into the enhanced dataset.

Looking ahead, there are plenty of opportunities to expand and refine this approach. Strategies to cut down on computational costs, like selective augmentation, task-specific fine-tuning, or using smaller models, could make it easier to scale for large datasets or various bias categories. Broadening the method to tackle a wider array of biases—such as those related to race, religion, age, or disability—and assessing its effectiveness across different NLP tasks would help demonstrate its overall applicability. Merging LGSA with fairness-focused learning objectives, like counterfactual fairness or equalized odds, could further boost bias mitigation efforts. Plus, fine-tuning language models on balanced datasets might allow for the creation of safer and more representative synthetic data with minimal manual effort.

## IX. Conclusion

This paper tackles the ethical issue of algorithmic bias in text-based AI by introducing and validating a practical solution called LLM-Guided Synthetic Augmentation (LGSA). The goal of the study was to address representational imbalances and performance gaps among different subgroups in a controlled NLP classification task, all while keeping or enhancing overall effectiveness.

Our experiments show impressive improvements. The baseline classifier reached an overall accuracy of 96.7%, but it had a gender bias gap of 7.2%. A simple pronoun-swap augmentation managed to shrink that gap to just 0.7%, although it did drop the accuracy to 95.6%. On the other hand, LGSA boosted the overall accuracy to 99.1% and cut the bias gap down to 1.9%, all while maintaining label fidelity and fluency in human evaluations. These findings suggest that using semantically coherent, context-aware synthetic examples allows models to focus on relevant task signals instead of misleading attribute cues.

The importance of these findings is twofold. First, LGSA offers a transparent pre-processing framework that can enhance the balance between fairness and utility compared to basic balancing methods, although putting it into practice might need careful management of computational resources. Second, it lessens the need to gather more sensitive data, providing some lightweight privacy benefits. However, one drawback of this approach is that it relies heavily on the quality of the seed corpus and the generative model, which could carry biases and impact the effectiveness of the augmentation. Future research should look into ways to boost scalability, like selective augmentation, optimizing prompts, or fine-tuning smaller models for specific tasks. Moreover, LGSA should be tested across various protected attributes and intersectional subgroups, integrated with formal privacy measures (like DP-aware distillation), and combined with fairness objectives during processing to achieve a well-rounded optimization. By tackling these challenges, LGSA paves the way for fairer, more reliable, and widely applicable NLP systems.